\documentclass[conference,letterpaper]{IEEEtran}
\IEEEoverridecommandlockouts
\usepackage{cite}
\usepackage{amsmath,amssymb,amsfonts}
\usepackage{algorithm}
\usepackage{algpseudocode}
\algnewcommand{\Input}[1]{\State \textbf{Input:} #1}
\algnewcommand{\Output}[1]{\State \textbf{Output:} #1}
\usepackage{subcaption} 
\usepackage{graphicx}
\usepackage{enumitem}
\usepackage{textcomp}
\usepackage{xcolor}
\def\BibTeX{{\rm B\kern-.05em{\sc i\kern-.025em b}\kern-.08em
    T\kern-.1667em\lower.7ex\hbox{E}\kern-.125emX}}
\begin{document}

\title{Think Fast: Real-Time Kinodynamic Belief-Space Planning for Projectile Interception}

\IEEEoverridecommandlockouts
\author{
    \IEEEauthorblockN{Gabriel Olin, Lu Chen, Nayesha Gandotra, Maxim Likhachev, Howie Choset}
    \IEEEauthorblockA{
        Carnegie Mellon University, Pittsburgh, PA, USA
    }
}
\addtolength{\topmargin}{0.2in}   % increase until the title drops into the margin box
\IEEEaftertitletext{\vspace{-0.5cm}}
\maketitle
\begin{abstract}
Intercepting fast moving objects, by its very nature, is challenging because of its tight time constraints. This problem becomes further complicated in the presence of sensor noise because noisy sensors provide, at best, incomplete information, which results in a distribution over target states to be intercepted. Since time is of the essence, to hit the target, the planner must begin directing the interceptor, in this case a robot arm, while still receiving information. We introduce an tree-like structure, which is grown using kinodynamic motion primitives in state-time space. This tree-like structure encodes reachability to multiple goals from a single origin, while enabling real-time value updates as the target belief evolves and seamless transitions between goals. We evaluate our framework on an interception task on a 6 DOF industrial arm (ABB IRB-1600) with an onboard stereo camera (ZED 2i). A robust Innovation-based Adaptive Estimation Adaptive Kalman Filter (RIAE-AKF) is used to track the target and perform belief updates.
\end{abstract}
\vspace{-0.33\baselineskip}
\vspace{-0.4cm}
\section{Introduction}\vspace{-0.3cm}
\vspace{-0.1cm}
Intercepting fast-moving objects requires split-second decisions under incomplete information. While a robot must begin moving toward an intercept even before the object’s trajectory is fully observed, sensors provide noisy and partial estimates, producing a range of possible future states. This makes the interception problem not only a matter of planning dynamically feasible motions but also of reasoning under uncertainty in real time.

Prior work in robotic interception has shown how dynamics and timing can be incorporated into planners for interception \cite{Bauml2010, Sintov2014}, but these methods often have a narrow success rate when dealing with noisy sensing, as they do not reason about uncertainty. In other motion planing problems with moving obstacles, such as autonomous driving, \cite{Aoude2010, Aoude2013, Bai2015} planners explicitly reason about evolving beliefs of moving agents to plan robust collision-avoiding trajectories, but often these planners can only run at several Hz. For close range interception, where planning and execution need to be done within a fraction of a second, it becomes necessary to develop even more efficient algorithms. 

In this work, we introduce a tree-based kinodynamic planner in state–time space, constructed from motion primitives rather than extend operators. This structure compactly encodes reachability to multiple goals, supports fast value updates as the target belief evolves, and enables seamless switching between intercept candidates. We combine this planner with an Innovation-based Adaptive Estimation Adaptive Kalman Filter (IAE-AKF) to maintain a belief over target trajectories. We evaluate the approach with hardware experiments on a 6-DOF ABB IRB-1600 robot arm and a ZED 2i stereo camera, demonstrating robust interception under uncertainty.
\section{Related Work}

\subsection{Projectile Interception and Robotic Catching}
Recent efforts on projectile interception with robot manipulators have explored both optimization-based planning and learning-based approaches. Bäuml et al.~\cite{Bauml2010} present one of the earliest demonstrations of kinematically optimal ball catching with a 7-DOF arm and 12-DOF hand. Their approach leverages stereo vision and an extended Kalman filter (EKF) for trajectory prediction, while solving for catch time and configuration as a nonlinear constrained optimization problem. Real-time feasibility is achieved via parallel SQP, supporting multiple catching behaviors. 

Sintov and Shapiro~\cite{Sintov2014} propose the Time-Based RRT (TB-RRT), which augments RRT with temporal information to enable rendezvous planning under timing constraints. The algorithm is validated on interception tasks such as a 1-DOF bat striking a ball and a 3-DOF robotic arm catching a moving object, demonstrating precise time-synchronized planning. 

Natarajan et al.~\cite{Natarajan2024} propose a preprocessing-based kinodynamic motion planning framework that leverages the INSAT planner and dome discretization to precompute smooth trajectories offline, enabling online retrieval of collision-free intercept motions in under a millisecond. While effective under strict real-time constraints, this approach relies on open-loop planning, which only uses one trajectory prediction to plan, and does not reason about uncertainty.

Kim et al.~\cite{Kim2014} present a programming-by-demonstration method that learns models of object dynamics and grasping configurations, using dynamical systems to generate rapidly adaptable catching motions. Their method achieves successful catching of diverse non-spherical objects but requires extensive demonstration data and motion capture infrastructure to train reliable models. Although these approaches consider dynamics and time constraints in the planning problem, they do not explicitly reason about uncertainty.

\subsection{Belief-Space Planning under Uncertainty}
In certain planning under uncertainty problems, the belief evolution of objects in the robot's environment is decoupled from action selection, which enables efficient planning approximations. In the autonomous driving space, the problem of online belief space planning has been applied to driving scenarios with uncertain obstacle motion or pedestrian intention. Aoude et al.~\cite{Aoude2010} introduce RRT-Reach, a sampling-based threat assessment framework for driver assistance at intersections. Coupled with a Bayesian intention predictor, RRT-Reach evaluates escape trajectories in real time to recommend evasive maneuvers. Their Threat Assessment Module (TAM) is validated in both autonomous and human-driven vehicle experiments, showing effective detection and avoidance of intersection violations. Aoude et al.~\cite{Aoude2013} also extend this work to handle dynamic obstacles with uncertain motion patterns, combining Chance-Constrained RRT (CC-RRT) with Regression-Reduced Gaussian Processes (RR-GP). This approach enforces per-step probabilistic safety guarantees under Gaussian uncertainty and outperforms naive and nominal planners in real-time simulations. 

Bai et al.~\cite{Bai2015} apply POMDP planning to autonomous driving in crowded environments, explicitly reasoning about pedestrian intentions. Their intention-aware online planner achieves real-time performance (3 Hz) and demonstrates robustness in dynamic multi-agent scenarios. 

\section{Problem Formulation}
Consider a small projectile that is launched towards a vertical plane 0.8 meters from the base of fixed-based robot manipulator. The state of the projectile is represented as $\rho = (\rho_p, \rho_v)$, where $\rho_p \in \mathbb{R}^3$ denotes its position and $\rho_v \in \mathbb{R}^3$ its velocity. As noisy sensor readings describing the position of the projectile come in, the manipulator must move its end effector to block it before it passes the plane, which covers the region we want to protect.
\begin{figure}[ht]
    \centering
        \centering
        \includegraphics[width=\linewidth]{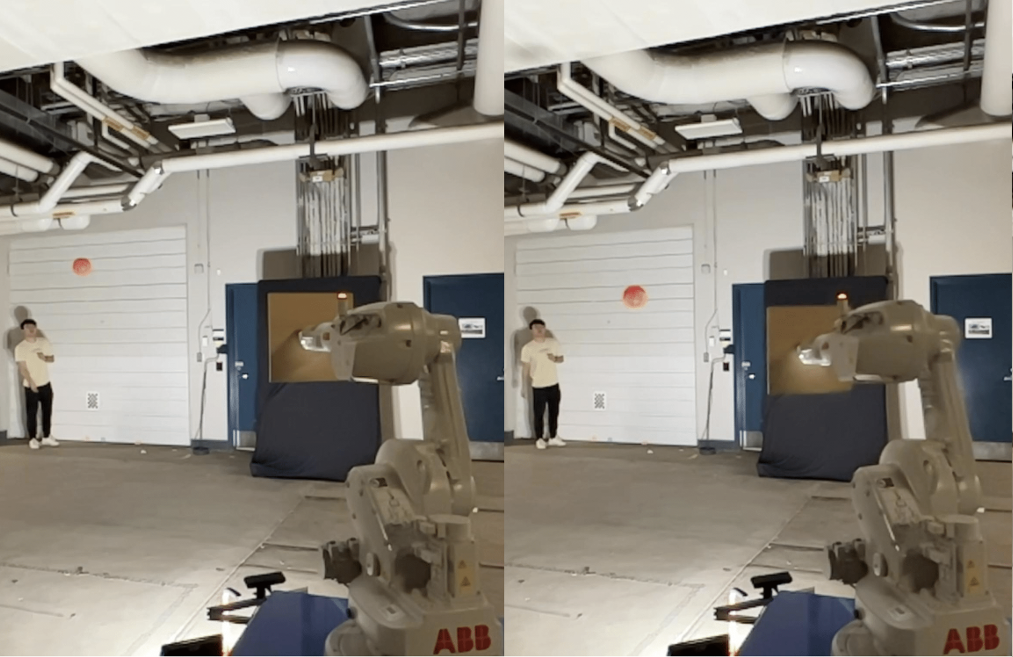}
        \caption{ABB arm with a shield attached to its end-effector blocking a thrown projectile}
        \label{fig:system}
\end{figure}
\begin{figure}[ht]
    \centering
        \centering
        \includegraphics[width=0.9\linewidth]{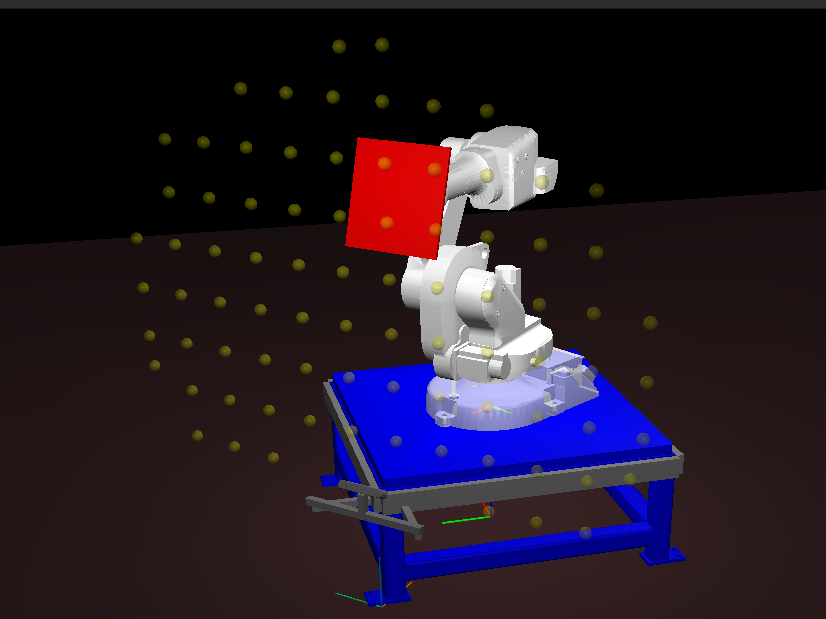}
        \caption{Task Space Goal Poses}
        \label{fig:IK}
\end{figure}

\subsection{System Overview}
Our manipulator is an ABB IRB-1600 robot arm with six DOF velocity control equipped with a rigidly attached shield to its end effector (Fig.~\ref{fig:system}). To track and infer the state of the projectile, a ZED 2i stereo camera is attached to the base of the robot which gives regular RGB-D readings at 50Hz, which can be segmented into timestamped \((x, y, z)\) coordinates. The intercept plane is chosen to be orthogonal to the camera's direction of view. To make the problem tractable in real time, we introduce the following simplifying assumptions:  
\begin{itemize}
    \item The manipulator always starts execution from a fixed ``home'' configuration $q_{\text{home}}$.  
    \item Projectiles are launched from within the field of view of the onboard stereo camera, ensuring early state estimates of their trajectories.  
    \item At any given time, only a single projectile is present in the environment.  
\end{itemize}
\subsection{Goal States}
To cover the maximum amount of the robot's workspace, we compute inverse kinematics solutions for the end effecetor for \((y, z)\) coordinates at a 0.75 meter discretization in the plane frame (Fig.~\ref{fig:IK}). The orientation of the end effector is constrained so that the shield is coplanar with the plane. The joint configurations computed by a numerical IK solver are saved as goal states for the planner.

\section{Computing Belief over Goal States}

We consider a stochastic projectile state under gravitational acceleration $g$ represented by a Gaussian belief, updated online by an Adaptive Kalman Filter (see Appendix)
\[
s = 
\begin{bmatrix}
X \\ Y \\ Z \\ V_x \\ V_y \\ V_z
\end{bmatrix}
\sim \mathcal{N}(\mu, \Sigma),
\]
where $(X,Y,Z)$ are positions and $(V_x,V_y,V_z)$ are velocities, all expressed in the plane frame. We are interested in the time $\tau$ at which the trajectory crosses the interception plane $x = x^*$, as well as the corresponding lateral coordinates $(Y,Z)$ at that crossing.

Although computing the probability distribution over the projectile's state and time when it crosses the intercept plane is generally computationally intractable due to its nonlinear dependence on $V_x$, the strict time budget of our problem necessitates a cheap analytical approximation to evaluate our belief over goal states online.
\subsection{First-order Delta Method}
The crossing time at the plane of interception in the robot's workspace is given by
\[
\tau = \frac{x^* - X}{V_x}.
\]
Using a first-order linearization about the mean state $\mu$, the distribution of $\tau$ is approximated as Gaussian:
\[
\mathbb{E}[\tau] \approx \bar{\tau} = \frac{x^* - \mu_X}{\mu_{V_x}}, 
\qquad
\mathrm{Var}[\tau] \approx g_\tau \, \Sigma \, g_\tau^\top,
\]
where the gradient of $\tau$ with respect to the state is
\[
g_\tau =
\begin{bmatrix}
-\tfrac{1}{V_x} & 0 & 0 & -\tfrac{x^*-X}{V_x^2} & 0 & 0
\end{bmatrix}.
\]
At the crossing time, the lateral coordinates are
\[
Y_{\text{plane}} = Y + V_y \,\tau, 
\qquad
Z_{\text{plane}} = Z + V_z \,\tau + \tfrac{1}{2} g \tau^2,
\]
where $g$ is the gravitational acceleration (with the sign chosen by convention). Defining the output vector:
\[
o = 
\begin{bmatrix}
Y_{\text{plane}} \\ Z_{\text{plane}} \\ \tau
\end{bmatrix},
\]
we introduce the mapping $h:\mathbb{R}^6 \to \mathbb{R}^3$ such that
\[
o = h(s),
\]
Because $h:\mathbb{R}^6 \to \mathbb{R}^3$ is a nonlinear mapping between the current state and the crossing state/time, the distribution of $o = h(s)$ is non-Gaussian. The \emph{delta method}\cite{ReithBraun2023ApproxFPTD} provides a first-order approximation by linearizing $h$ about the mean $\mu$:  
\[
h(s) \;\approx\; h(\mu) + J (s - \mu),
\]
where 
\[
J = \frac{\partial h}{\partial s}\Big|_{s=\mu}
\]
is the Jacobian of $h$ evaluated at $\mu$.  
The rows of $J \in \mathbb{R}^{3\times 6}$ are given by the partial derivatives of $(Y_{\text{plane}},Z_{\text{plane}},\tau)$: \[ \frac{\partial Y_{\text{plane}}}{\partial s} = \begin{bmatrix} -\tfrac{V_y}{V_x} & 1 & 0 & -\tfrac{V_y \tau}{V_x} & \tau & 0 \end{bmatrix}, \] \[ \frac{\partial Z_{\text{plane}}}{\partial s} = \begin{bmatrix} -\tfrac{V_z}{V_x} - \tfrac{g \tau}{V_x} & 0 & 1 & -\tfrac{V_z \tau}{V_x} - \tfrac{g \tau^2}{V_x} & 0 & \tau \end{bmatrix}, \] \[ \frac{\partial \tau}{\partial s} = \begin{bmatrix} -\tfrac{1}{V_x} & 0 & 0 & -\tfrac{x^*-X}{V_x^2} & 0 & 0 \end{bmatrix}. \]
Under this linear approximation, $o$ is approximately Gaussian:
\[
o \sim \mathcal{N}\!\big(h(\mu), \; J \Sigma J^\top\big).
\]
We apply this construction with $h(s)$ defined as the mapping from the projectile state $s = (X,Y,Z,V_x,V_y,V_z)$ to the lateral intercept coordinates and crossing time $o = (Y_{\text{plane}}, Z_{\text{plane}}, \tau)$.

\subsection{Joint Gaussian at Plane}
This construction yields a joint Gaussian approximation for the lateral coordinates at plane crossing and the crossing time:
\[
\big(Y_{\text{plane}}, Z_{\text{plane}}, \tau\big)^\top \;\sim\; 
\mathcal{N}\big(\bar{o}, \Sigma_o\big),
\]
which captures not only the individual variances but also the correlations between $\tau$, $Y_{\text{plane}}$, and $Z_{\text{plane}}$ induced by the uncertain initial state.

\section{State-time Belief Space Planning}

\subsection{Action Tree}
Offline, we construct an \emph{action tree} rooted at the robot's initial state in $\mathbb{R}^9$, where each node encodes a kinematic state consisting of joint position, velocity and acceleration $(p,v,a)$. Even though the ABB robot arm has six degrees of freedom, we plan only for the first three degrees of freedom because they account more most of the task space positional movement, which simplifies the planning problem. The orientation is controlled separately by moving the last three joints to align the shield with the incoming angle of the projectile. This decoupling is possible because the last three joints are able to move far faster and thus are not the limiting factor in the time-optimal control problem to move from one configuration to another. 

From the start state, we iteratively expand the tree using minimum-time motion primitives (see Section~B), steering towards sub-goals that span a discretization of the robot’s reachable front workspace plane. Primitives that are in collision are discarded and cut off at a maximum duration $\epsilon$. Each expansion adds dynamically consistent primitives up to a fixed maximum depth, while storing cost (time-to-come) at each successor node. We choose the tree depth to limit the number of direction changes the robot makes while the projectile is in motion - more decision nodes would result in highly jerky motion and increased strain on the motors. Nodes at the maximum depth are then connected to nearby goal states with zero velocity.

This procedure allows the tree to cover the state-time space between the start configuration and goal configurations, allowing the robot to consider paths during execution that maximize expected reachability to each of the goals (see Section C). The pseudocode for tree construction is shown in Algorithm~\ref{alg:actiontree}. An example action tree for a 2D double integrator is shown in Fig. 3.

\begin{algorithm}[t]
\caption{Action Tree Construction}
\label{alg:actiontree}
\begin{algorithmic}[1]
\Input{Start state $s_0$, goal states $G$, goal regions $\mathcal{R}$, state limits $\mathcal{X}$, control limits $\mathcal{U}$, horizon $\epsilon$, maximum depth $d_{\max}$}
\Output{Action tree $\mathcal{T}$}
\State Initialize tree $\mathcal{T}$ with $s_0$
\State Initialize queue $q \gets \{s_0\}$
\While{$\mathcal{T}.\text{max\_level} < d_{\max}$ \textbf{and} $q \neq \emptyset$}
  \State $s \gets q.\text{pop}$
  \If{$s.\text{level} = d_{\max}$}
    \State \Call{ConnectToCloseGoals}{$s, G$}
    \State \textbf{continue}
  \EndIf
  \ForAll{$g \in$ \Call{SubGoals}{$s, \mathcal{R}$}}
    \State \Call{$\mathcal{T}.\text{steer}$}{$s, g, \epsilon$} \Comment{$\epsilon$ is max duration}
    \State $q.\text{push}(g)$
  \EndFor
\EndWhile
\end{algorithmic}
\end{algorithm}

\begin{figure}[ht]
    \centering
        \centering
        \includegraphics[width=\linewidth]{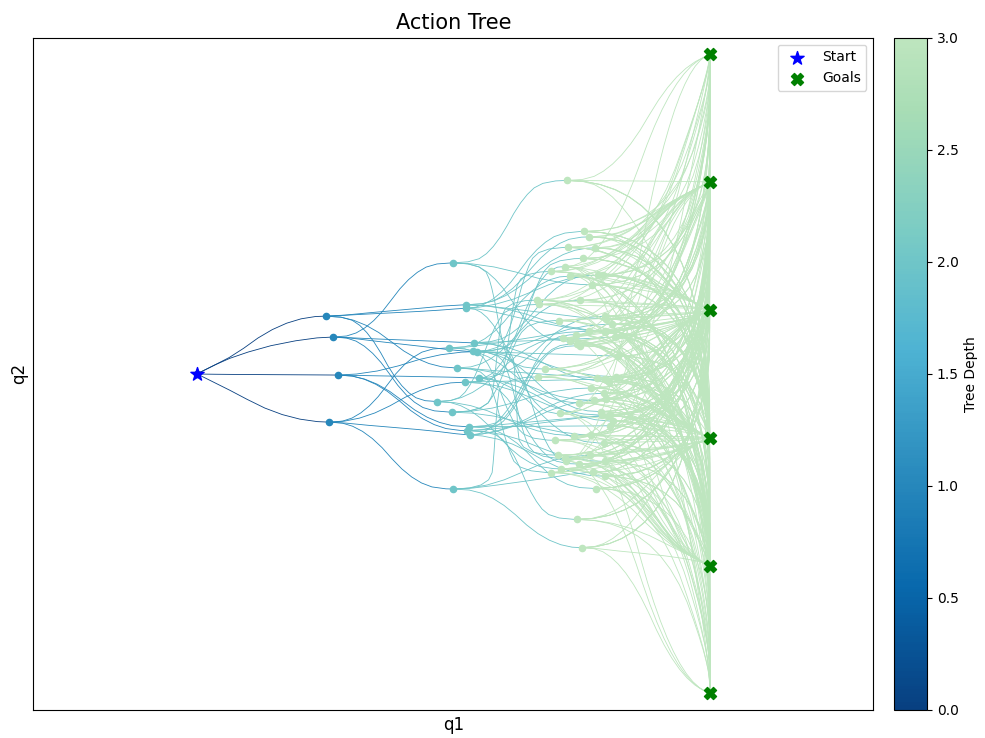}
        \caption{Action Tree for 2D Double Integrator}
        \label{fig:action_tree}
\end{figure}
\subsection{Minimum-time Optimal Control}
Kinodynamic planning requires satisfaction of both kinematic constraints (e.g., position and velocity limits) and dynamic constraints (e.g., actuation, force, or torque bounds). However, for fully actuated robotic manipulators, exact dynamic feasibility is often unnecessary because low-level joint controllers can accurately track reference trajectories as long as they respect kinematic limits. For this reason, we approximate each joint’s dynamics as a \emph{triple integrator system}:
\[
\dot{p} = v, \quad \dot{v} = a, \quad \dot{a} = u,
\]
where $u$ is the jerk control input.  

The objective is to compute the minimum-time, jerk-limited trajectory connecting an initial state $(p_0,v_0,a_0)$ to a terminal state $(p_f,v_f,a_f)$ subject to state and control bounds. The problem can be formulated as:
\[
\begin{aligned}
\min_{u(\cdot),\,T} \quad & J = T \\
\text{s.t.} \quad & \dot{a}(t) = u(t), \quad |u(t)| \leq u_{\max}, \\
& p_{\min} \leq p(t) \leq p_{\max}, \\
& v_{\min} \leq v(t) \leq v_{\max}, \\
& a_{\min} \leq a(t) \leq a_{\max}, \\
& p(0) = p_0, \quad v(0) = v_0, \quad a(0) = a_0, \\
& p(T) = p_f, \quad v(T) = v_f, \quad a(T) = a_f. \\
\end{aligned}
\]
From Pontryagin's Minimum Principle, this formulation admits a bang–off–bang profile for the optimal control $u(t)$, which can be solved in closed form or via efficient numerical methods, providing minimum-time steering primitives for the action tree. We use the highly optimized Ruckig implementation \cite{berscheid2021jerk} as our solver, because it can handle higher-order non-zero terminal constraints. 

\subsection{Value Function}
To evaluate the quality of a candidate action, we define a \emph{value function} over the state-time belief space. Each \emph{terminal node} in the tree corresponds to a unique sequence of motion primitives that reaches a particular goal state $g \in \mathcal{G}$ with an associated arrival time $\tau_g$. 

Let the belief over the lateral intercept coordinates and crossing time be represented by a Gaussian distribution
\[
b(o) \;=\; \mathcal{N}\!\big(o \,\big|\, \bar{o}, \Sigma_o\big),
\qquad
o = \begin{bmatrix} Y_{\text{plane}} \\ Z_{\text{plane}} \\ \tau \end{bmatrix}.
\]
Because our camera is fixed to the base frame, our actions are not coupled to the observations we receive. This allows for the use of the QMDP approximation of the underlying POMDP, where our actions are the motion primitives in the tree, the beliefs are the Gaussian distribution over lateral intercept coordinates and time, and reward is a binary signal of whether the shield blocks the projectile before intercept time. Under the QMDP approximation, the value of a node $n$ under belief $b(o)$ can be written as
\[
V(n, b(o)) 
\;=\; \mathbb{E}_{o \sim b}\!\Big[ R(n,o) \;+\; \max_{a \in \mathcal{A}} \; V(n', b(o)) \Big],
\]
where $R(n,o)$ is the reward of being at node $n$ 

We define the reward function only for terminal nodes. For a terminal node $n$, let $S$ denote the set of lateral coordinates $(Y,Z)$ covered by the shield at the intercept plane. Then
\[
R(n,o) \;=\;
\begin{cases}
1, & (o_Y, o_Z) \in S \;\;\text{and}\;\; n.\text{time} < o_\tau, \\[1ex]
0, & \text{otherwise},
\end{cases}
\]
where $o = (o_Y, o_Z, o_\tau)$ are the intercept coordinates and the crossing time. This value function represents the expected probability of successfully intercepting the projectile, given the current distribution over intercept coordinates and crossing time.

The tree structure of our decision process allows terminal node values to be computed with each belief update and then propagated backward through the tree in a single online pass. During execution, the next primitive is selected by choosing the successor of the current node that maximizes the value function:
\[
a^* \;=\; \arg\max_{a \in \mathcal{A}(n)} V(n', b),
\]
where $\mathcal{A}(n)$ denotes the set of available successors of node $n$, and $n'$ is the child node reached by applying action $a$. If a terminal node is reached while projectile measurements are still being received, the planner tells the robot to move towards the most probable goal.

The closed-form Gaussian approximation of the belief state, along with efficient value updates in our action tree, allows for online planning. Our planner consistently computes the next action online in under 10ms on an AMD Threadripper Pro 5995WX workstation.

\section{Experiments}
We tested the full system with integrated planning, perception, and control on an IRB-1600 robot arm, equipped with a ZED 2i stereo camera in an indoor environment. Balls were thrown towards the robot's front plane from distances from 6-8m. The time of flight of the throws were typically between 400ms and 800ms. We compared our method to a naive baseline, in which the robot moves (under kinematic constraints) as fast as possible to the current most likely goal configuration at each belief update. 

Out of 50 throws of varying initial velocity and distance, our method achieved a blocking success rate of 74\%. The naive approach still achieved 63\% blocking success. The main cause of failure was insufficient execution time. While the perception module eventually returned an accurate intercept location, the end effector was too far away to reach it in time under its kinematic limits.

\section{Discussion}
While both our method and the naive approach did not achieve 100\% success rate, the improvement shows a benefit to considering uncertainty in extreme time-constrained interception problems. By encoding time-constrained reachability in the value function, our planner is able to hedge against a distribution of future observations, making it more robust than the naive approach which commits immediately to the first information it receives. Further work will include optimizing for manipulability of the robot arm in the motion primitives, which may allow for more agility when needing  rapidly changing directions in task space. Additionally, the discrete action space may be limiting to when and how the robot can move, so optimization or learning based approached may provide a strong continuous action space analog to our method.

\section{Conclusion}
We presented a real-time kinodynamic belief space planning framework for robotic projectile interception under uncertainty. By constructing an action tree in state–time space with jerk-limited motion primitives, our method enables fast online value updates as the projectile belief evolves, allowing the robot to seamlessly adapt its interception strategy. The QMDP-based value function, defined over the Gaussian distribution of intercept states, provides a principled approximation to reason about success probabilities under partial information.  

Our integrated system—combining robust RIAE-AKF state estimation with kinodynamic motion primitives—demonstrated improved interception success over a naive baseline on  hardware experiments with a 6-DOF ABB IRB-1600 manipulator. Despite inherent time constraints and occasional failures due to execution limits, the results highlight the importance of explicitly modeling uncertainty close range interception tasks.  

Future work will explore incorporating robot manipulability into the value function, expanding from discrete primitives to continuous optimization or learning-based action spaces, and extending the framework to multi-interceptor scenarios. These directions could further enhance agility and robustness in interception tasks, bringing robotic systems closer to human-level performance in highly dynamic environments.

\bibliographystyle{IEEEtran}   % or another style
\bibliography{references}
\appendix
\subsection*{Perception and State Estimation}\label{A}
\begin{figure}[ht]
    \centering
        \centering
        \includegraphics[width=\linewidth]{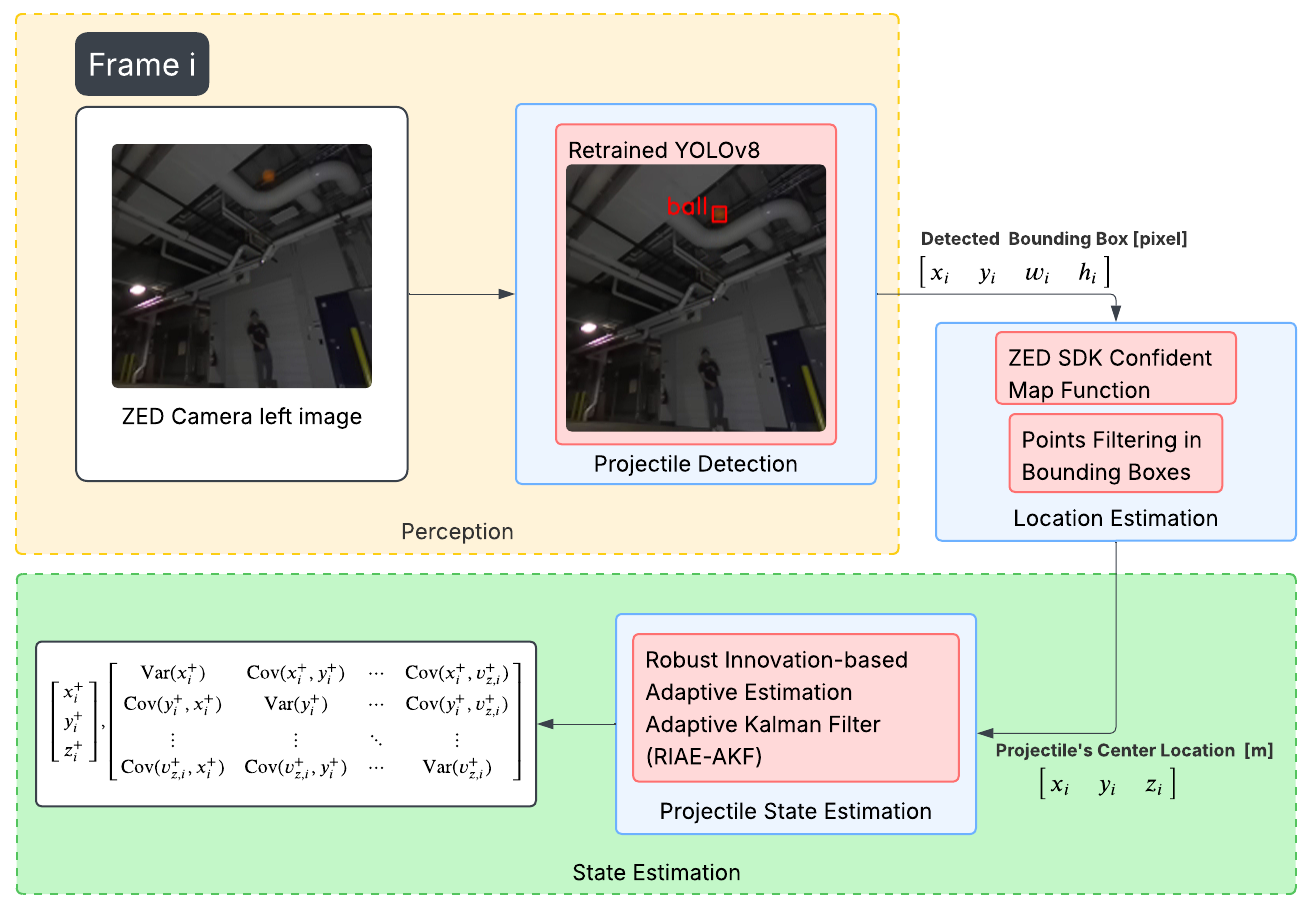}
        \caption{Flowchart showing Perception and State Estimation}
        \label{fig:perception}
\end{figure}
Perception is affected by multiple sources of noise, including low-resolution images, lighting variations, and complex backgrounds. Simple detection algorithms like a color filter, as used in ~\cite{Natarajan2024},  can struggle with robustness to these conditions. To reduce these effects and improve detection accuracy, we use a retrained YOLOv8 (You Only Look Once, version 8)~\cite{yolov8_ultralytics} to detect the projectile in each frame using laboratory videos. The DINO grounding~\cite{liu2023} was used as the base model, with AutoDistill CaptionOntology used to generate the training dataset. A custom annotation GUI was developed, allowing human verification and refinement of the ground truth. For each frame from the left ZED camera, the perception module detects the projectile and outputs its bounding boxes \([x_i, y_i, w_i, h_i]\), which contain the center point in the x and y axis and its width and height.
To obtain the 3D location of the projectile, the detected bounding boxes are adjusted using a reverse sigmoid function based on the ratio of the bounding box area and image area. It keeps shrinking as the ball approaches the camera: 5.3\% shrinkage in width and length when the ball is about 7-8 meters away, and up to 27.6\% shrinkage when it is very close. These modified bounding boxes are then used to retrieve the projectile locations from ZED SDK’s confidence map and point clouds~\cite{Natarajan2024}. This adjustment reduces the noise from motion blur when the ball moves rapidly near the camera. The module outputs the projectile's estimated 3D center location \([x_i, y_i, z_i]\).

A total of 1,408 images were used for training, including both high- and low-resolution cases, as well as scenarios in which the ball was held in the hand. The proposed perception module can detect the ball earlier than the color filter method (that is, before it is thrown), achieving an 81.3\% detection success rate across 385 evaluation images, with an average center difference of 0.97 pixel units compared to the ground-truth bounding box center. In contrast, the color filter method achieved only a success rate of 38. 1\% and a difference of 1.94 pixel units. An overview of the perception and state estimation pipeline is given in Fig. 2.

\begin{figure}[ht]
    \centering
    % First row
    \begin{subfigure}{0.45\linewidth}
        \centering
        \includegraphics[width=\linewidth]{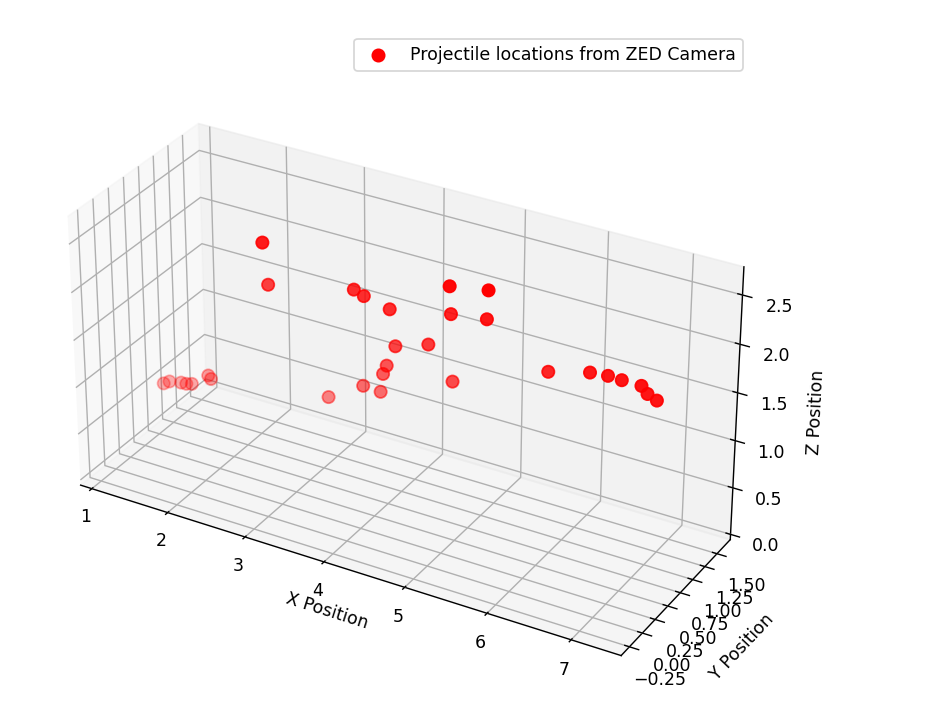}
    \end{subfigure}
    \begin{subfigure}{0.45\linewidth}
        \centering
        \includegraphics[width=\linewidth]{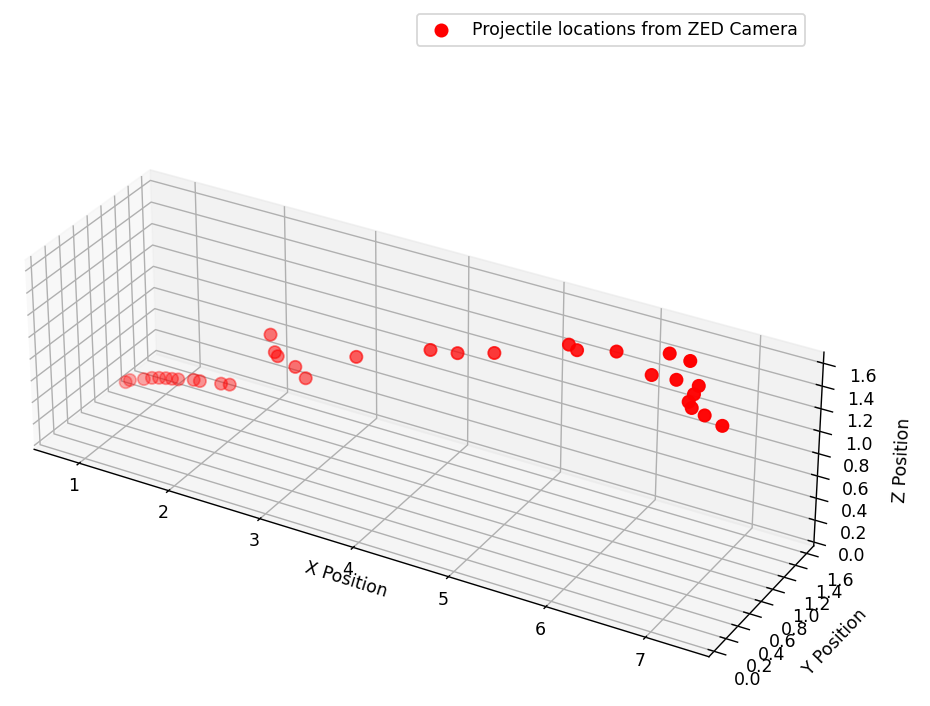}
    \end{subfigure}
    \caption{The left and right figures show 3D locations of the projectile at different throws. Measurement noise varies between throws.}
    \label{fig:measurement noises}
\end{figure}

While detecting and segmentating the position of a projectile at several timesteps, a Kalman Filter (KF) can optimally estimate its state accounts for measurement and process noise with covariance matrices Q and R, but only when both are constant. In our case, different throws produce varying measurement noise (Fig.~\ref{fig:measurement noises}), which makes the filtering process difficult. A constant process noise covariance is suitable since the model is only influenced by gravity $(g = 9.81\,\text{m/s}^2)$. The innovation-based adaptive estimate adaptive Kalman filter (IAE-AKF) algorithm adjusts Q and R using the innovation history. However, when there is disturbed noise or outliers, its residual can exceed Gaussian assumptions, leading to a degradation of performance~\cite{Xian2013}. To consider the varying measurement noise covariance and in the projectile’s locations, we use Robust IAE-AKF (RIAE-AKF)~\cite{Xian2013}~\cite{Silva2022} 

Similar to the Kalman Filter, we need to calculate raw innovation \(\boldsymbol{\nu}_k\) and covariance matrix after getting standard kalman prediction \(\hat{\mathbf{x}}_k^-\)  and \(\mathbf{P}_k^-\).
\begin{equation}
\boldsymbol{\nu}_k = \mathbf{z}_k - \mathbf{H}_k \hat{\mathbf{x}}_k^- \label{eq:innovation}
\end{equation}
Where \(\mathbf{H}_k\) is the observation matrix. For each element, calculate the chi-square statistic \(\kappa_k(i)\)\cite{Geng2008}:

\begin{equation}
\kappa_k(i) = \frac{\nu_k^2(i)}{\mathbf{C}_{v_{k-1}}(i,i)} \sim \chi^2(1) \label{eq:kappa_i}
\end{equation}
\(\nu_k(i)\) is the i-th element of the vector \(\nu_k\), \(\mathbf{C}_{v_{k-1}}(i,i)\) is the i-th diagonal element of empirical innovation covariance matrix \(\mathbf{C}_{v_{k-1}}\), shown in Eq.~\eqref{eq:Cv}. \(\chi^2(m)\) is a Chi-square distribution with m degree of freedom, and m is the number of innovation vector’s elements which we define as 1, assuming each dimension is independent to the other. 

Next, we revise the innovation to reduce the influence of abnormal innovation vectors and update its covariance using the innovation history. Here, \(\alpha\) represents the confidence level. The revised innovation is computed as
\begin{equation}
\hat{\nu}_k(i) = \begin{cases}
\nu_k(i), & \text{if } 0 \leq \kappa_k(i) < \chi_\alpha^2(1) \\[0.2em]
\nu_k(i) \exp\left(-\frac{\kappa_k(i) - \chi_\alpha^2(1)}{\chi_\alpha^2(1)}\right), & \text{if } \kappa_k(i) \geq \chi_\alpha^2(1)
\end{cases} \label{eq:chi}
\end{equation}
When \(0 \leq \kappa_k(i) < \chi_\alpha^2(1)\), the sensor reading is consistent with expectations. If \(\kappa_k(i) \geq \chi_\alpha^2(1)\), the innovation is abnormal and will need to be modified using Eq.~\eqref{eq:chi}. Update Innovation covariance matrix \(\mathbf{C}_{v,k}\):
\begin{equation}
\mathbf{C}_{v,k} = \frac{1}{N} \sum_{j=k-N+1}^{k} \hat{\boldsymbol{\nu}}_j \hat{\boldsymbol{\nu}}_j^\top \label{eq:Cv}
\end{equation} $N$ is the window size and $\hat{\boldsymbol{\nu}}_j$ are the revised innovation vectors. Then we update the measurement noise covariance matrix \(\mathbf{R}_k\).
\begin{equation}
\mathbf{R}_k = \mathbf{C}_{v,k} - \mathbf{H}_k \mathbf{P}_k^- \mathbf{H}_k^\top 
\end{equation}
The State Estimation module adapts the measurement noise, recursively updates the state, and outputs the projectile state 
\(\mathbf{x}_i^+ = [x_i^+, y_i^+, z_i^+, v_{x,i}^+, v_{y,i}^+, v_{z,i}^+]^T\) 
and the covariance matrix \(P\):
\[
P = 
\begin{bmatrix}
\operatorname{Var}(x_i^+) & \operatorname{Cov}(x_i^+, y_i^+) & \cdots & \operatorname{Cov}(x_i^+, v_{z,i}^+) \\
\operatorname{Cov}(y_i^+, x_i^+) & \operatorname{Var}(y_i^+) & \cdots & \operatorname{Cov}(y_i^+, v_{z,i}^+) \\
\vdots & \vdots & \ddots & \vdots \\
\operatorname{Cov}(v_{z,i}^+, x_i^+) & \operatorname{Cov}(v_{z,i}^+, y_i^+) & \cdots & \operatorname{Var}(v_{z,i}^+)
\end{bmatrix}.
\]
\vspace{-0.0cm} % adjust this value as needed
\begin{figure}[t]
    \centering
    \begin{subfigure}[t]{0.45\linewidth}
        \centering
        \includegraphics[width=\linewidth]{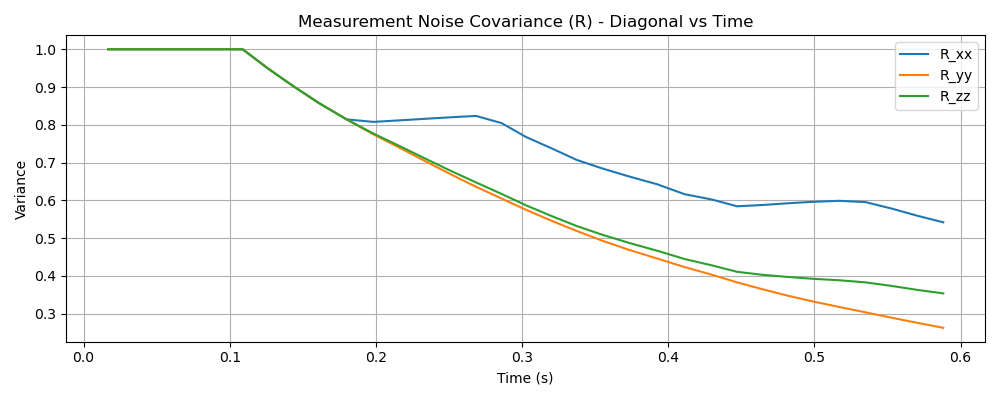}
        \caption{Measurement Noise Covariance Matrix $R$}
    \end{subfigure}
    \hfill
    \begin{subfigure}[t]{0.45\linewidth}
        \centering
        \includegraphics[width=\linewidth]{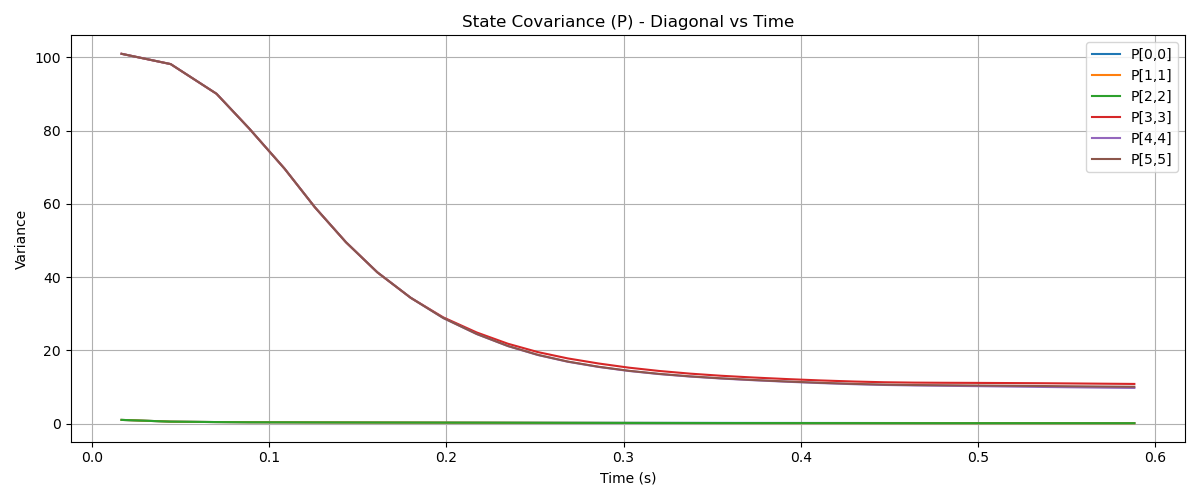}
        \caption{State Covariance Matrix $P$}
    \end{subfigure}
    \caption{Measurement noise covariance matrix $R$ and state covariance matrix $P$ over time for the throw shown on the left in Fig.~\ref{fig:measurement noises}. The diagonal values of R vary due to inaccurate locations in the middle of the throw; the diagonal values of $P$ decrease as expected when more information is given.}
    \label{fig:RP}
\end{figure}

\begin{figure}[ht]
    \centering
    \begin{subfigure}[t]{0.48\linewidth}  
        \centering
        \includegraphics[width=\linewidth]{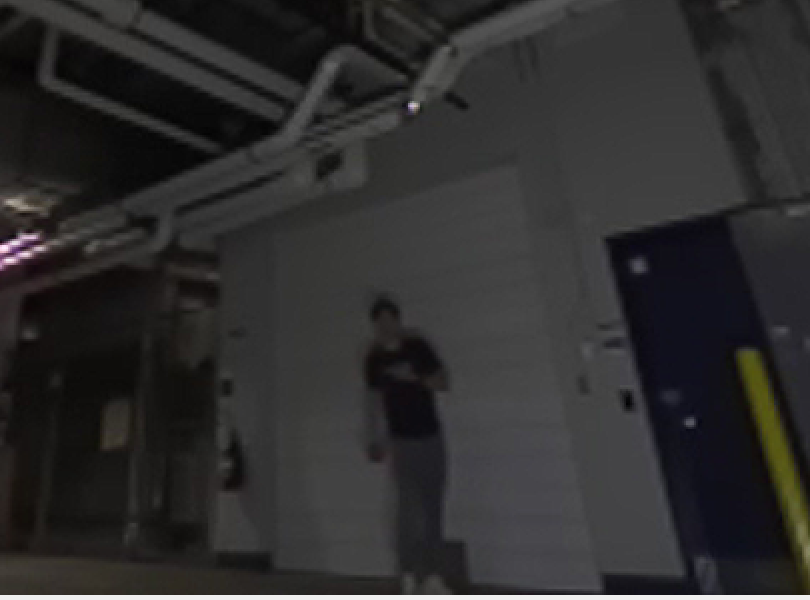}
        \caption{An example of a zoomed in frame received by the left ZED camera.}
    \end{subfigure}
    \hfill 
    \begin{subfigure}[t]{0.48\linewidth}
        \centering
        \includegraphics[width=\linewidth]{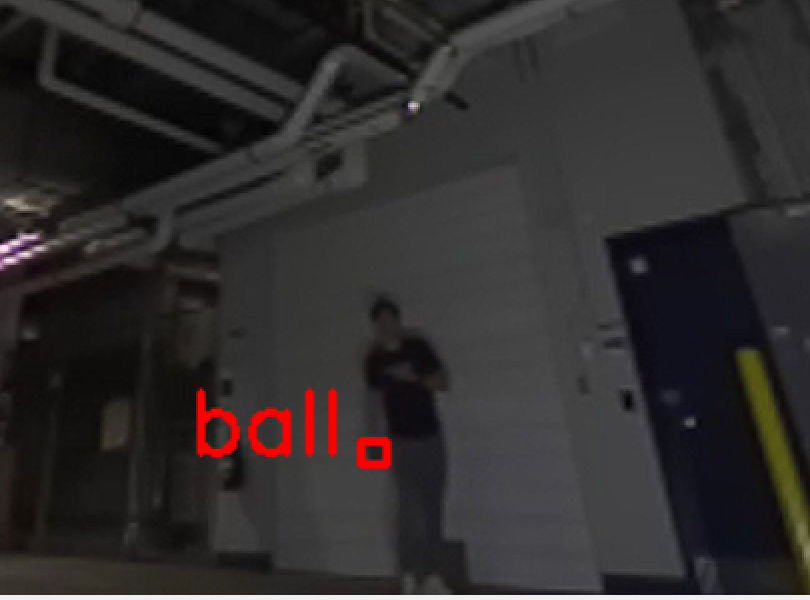}
        \caption{Same frame with the projectile detected, which the Color Filter fails.}
    \end{subfigure}
    \caption{Unlike the Color Filter~\cite{Natarajan2024}, the retrained YOLOv8 can detect the projectile in the early stage, even before the ball was thrown.}
    \label{fig:perception_images}
\end{figure}
To evaluate performance compared to the work~\cite{Natarajan2024}, given the absence of ground-truth 3D projectile locations, we used the estimated depth results to compute the intersection with a virtual vertical wall placed 2 meters from the camera. This intersection point serves as a reference point for ground truth, since the measurements are empirically closer to the real trajectories of the projectile. The intersection predicted by RIAE-AKF is then calculated, and its distance from the reference point of the ground truth is calculated to compare with the work~\cite{Natarajan2024}. shows that the work~\cite{Natarajan2024} has fluctuated prediction compared to the proposed work. 

\end{document}